%% file: main.tex
\documentclass[conference]{IEEEtran}
\IEEEoverridecommandlockouts
\usepackage{algorithm}
\usepackage{algorithmic}

\usepackage{cite}
\usepackage{amsmath,amssymb,amsfonts}
\usepackage{algorithmic}
\usepackage{graphicx}
\usepackage{textcomp}
\usepackage{xcolor}
\usepackage{subcaption}
\usepackage{multirow}
\usepackage{url}
\DeclareMathAlphabet{\mathbb}{U}{bbold}{m}{n}

\usepackage{eso-pic}
\usepackage{tikz}
\usepackage{textcomp}

\newcommand\copyrighttext{%
  \footnotesize \textcopyright 2026 IEEE. Personal use of this material is permitted.  Permission from IEEE must be obtained for all other uses, in any current or future media, including reprinting/republishing this material for advertising or promotional purposes, creating new collective works, for resale or redistribution to servers or lists, or reuse of any copyrighted component of this work in other works. 

  Accepted as a paper at the IEEE RAGE 2026 Workshop.}
\newcommand{\copyrightnotice}{%
\begin{tikzpicture}[remember picture,overlay]
\node[anchor=south,yshift=10pt] at (current page.south) {\fbox{\parbox{\dimexpr\textwidth-\fboxsep-\fboxrule\relax}{\copyrighttext}}};
\end{tikzpicture}%
}

\def\BibTeX{{\rm B\kern-.05em{\sc i\kern-.025em b}\kern-.08em
    T\kern-.1667em\lower.7ex\hbox{E}\kern-.125emX}}
\begin{document}

\title{
Hierarchical adaptive control for real-time dynamic inference at the edge}

\author{\IEEEauthorblockN{1\textsuperscript{st} Francesco Daghero}
\IEEEauthorblockA{\textit{MMMI Software Engineering} \\
\textit{University of Southern Denmark}\\
Odense, Denmark \\
fdag@mmmi.sdu.dk}
\and
{2\textsuperscript{nd} Mahyar Tourchi Moghaddam}\\
\IEEEauthorblockA{\textit{MMMI Software Engineering} \\
\textit{University of Southern Denmark}\\
Odense, Denmark \\
mtmo@mmmi.sdu.dk}
\and
{3\textsuperscript{rd} Mikkel Baun Kjærgaard}\\
\IEEEauthorblockA{\textit{MMMI Software Engineering} \\
\textit{University of Southern Denmark}\\
Odense, Denmark \\
mbkj@mmmi.sdu.dk}}

\maketitle
\copyrightnotice

\input{00_abstract}


\input{01_introduction}

\input{02_background}
\input{03_related}

\input{04_methodology}

\input{06_experimental_evaluation}
\input{conclusions}

\bibliographystyle{IEEEtran}
\bibliography{bib}

\end{document}

%% file: 00_abstract.tex
\begin{abstract}
Industrial systems increasingly depend on Machine Learning (ML), and operate on heterogeneous nodes that must satisfy tight latency, energy, and memory constraints.
Dynamic ML models, which reconfigure their computational footprint at runtime, promise high energy efficiency and lower average latency for modest accuracy tradeoffs; however, their deployment is complex due to the additional hyperparameters they rely on.
These hyperparameters, controlling the accuracy versus average latency tradeoff, are often tuned on a calibration dataset that must match the test time distribution, an assumption that rarely holds in real-world scenarios, leading to suboptimal operational conditions, possibly below static models.
We propose a two-tier adaptive architecture that co-optimizes model and system decisions.
At the global level, a constraint-driven scheduler configures and deploys, for each edge node, a cascade of classifiers composed of lightweight specialized models and a generalist fallback, satisfying latency and memory constraints.
At the node level, a fast local controller tracks data drifts and hardware resources, adaptively enabling or disabling specialized predictors (SP) to preserve high energy efficiency and avoid latency-constraint violations under varying conditions.
This design allows longer operating times without forcing a global redeployment step, and enables efficient execution in case of an unreachable remote global controller.
We evaluate the approach on two vision datasets under controlled distribution mismatch scenarios, showing average per-inference reductions of latency up to 2.45x and energy up to 2.86x, with less than 4\% accuracy drop compared to static baselines.
Our contributions are: (1) a budgeted SP-cascade formulation that preserves worst-case latency constraints; (2) a hierarchical controller that maintains efficiency under data and resource changes; and (3) an experimental evaluation on embedded hardware.
\end{abstract}

%% file: 01_introduction.tex
\section{Introduction}
Industrial systems increasingly rely on Machine Learning (ML) for real-time perception and control~\cite{automotive_industry,construction_machinery1}.
These tasks must satisfy strict latency constraints and are therefore often executed on edge nodes with intermittent or no cloud connectivity.
In this setting, systems must preserve timing predictability while remaining energy-efficient, since edge nodes are often battery-operated.
Dynamic inference techniques~\cite{adaptivereview} improve efficiency by adapting the computational graph depending on the input.

\begin{figure}[h]
    \centering
    \includegraphics[width=0.85\linewidth]{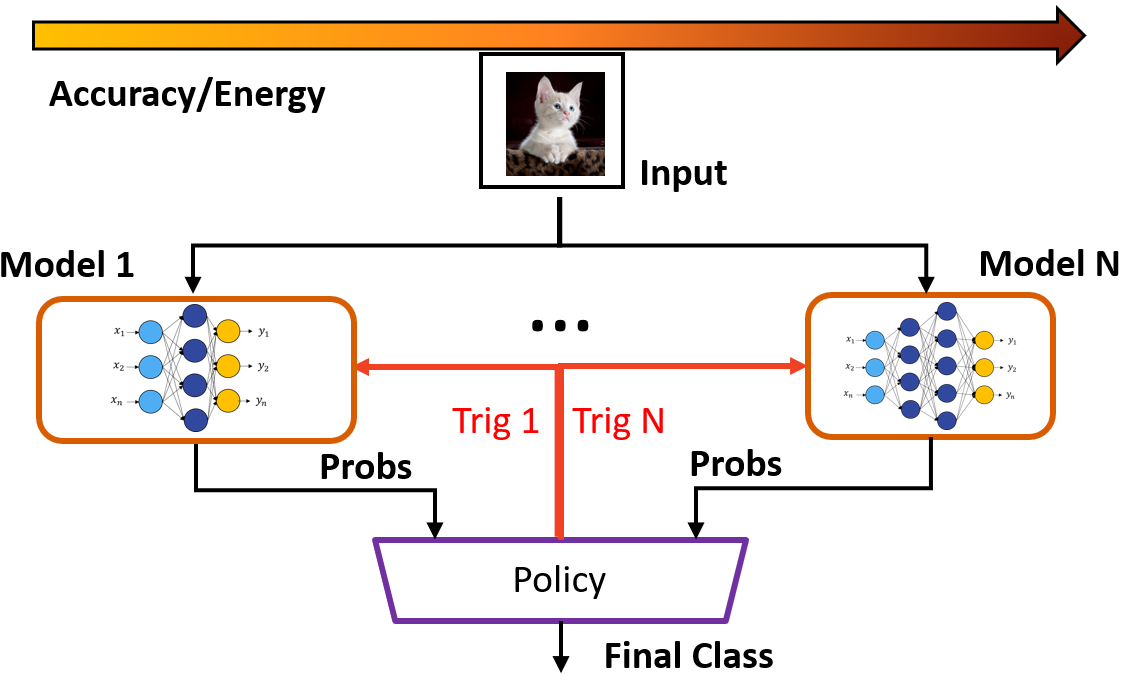}
    \caption{Generic dynamic inference scheme.}\vspace{-4mm}
    \label{fig:dynamic_inference_intro}
\end{figure}

As shown in Figure~\ref{fig:dynamic_inference_intro}, inputs traverse a cascade of models of increasing complexity, and a policy decides whether to execute the next model or exit.
Both the policy and the cascade are tuned on a validation set and assume the test distribution and the compute budget remain stable.
In practice, class distributions drift, and thermal throttling and background tasks reduce the available resources.
Dynamic inference can then become inefficient and may violate latency constraints if the selected cascade is too deep.
Recalibrating the cascade is time-consuming and can introduce unacceptable delays.

To address this challenge, we propose a two-tier self-adaptive software architecture for dynamic inference on heterogeneous edge systems.
At deployment time, a global scheduler selects, for each node and task, a cascade of inexpensive one-vs-rest binary classifiers, named specialized predictors (SPs), together with a multiclass fallback model (FB).
The number and order of SPs are optimized to minimize average latency while satisfying worst-case latency constraints.
At runtime, a local controller (LC) monitors resource usage and class frequencies to enable, disable, and reorder SPs to preserve efficiency.
A global controller (GC) runs at a lower frequency and can deploy additional SPs or update the configuration when needed, limiting the need for full redeployment.
This design decouples coarse-grained configuration from fine-grained model selection, enabling systems to remain responsive and energy-efficient.
Our work makes the following contributions:
\begin{itemize}
    \item A budgeted dynamic inference formulation based on SP cascades with an explicit worst-case latency constraint.
    \item A hierarchical controller that maintains efficiency under distribution mismatch and runtime resource variability.
    \item An evaluation on two vision datasets under controlled mismatch scenarios, showing average per-inference reductions of latency up to 2.45x and energy up to 2.86x with less than 4\% accuracy drop compared to static baselines.
\end{itemize}

%% file: 02_background.tex
\section{Background}~\label{sec:background}
ML, and in particular Deep Learning (DL), has become the core component of an increasing number of applications, such as remote healthcare~\cite{mlhealthcare}, and construction machinery~\cite{construction_machinery1}. 
However, the high computational demand of DL models has traditionally led to a cloud-centric paradigm, offloading inference to remote high-performance nodes.
While effective for training, cloud inference is often suboptimal~\cite{edgegood} for inference in many industrial scenarios due to strict real-time latency constraints and intermittent or absent connectivity.
On the other hand, near-sensor inference offers several advantages.
First, local computations ensure predictable and possibly lower average and worst-case latency~\cite{edgegood}.
Second, no sensitive data is transmitted to the cloud, enhancing security.
Finally, it can improve energy efficiency~\cite{edgegood}, as continuous transmission of raw sensor data to the cloud is highly expensive in terms of energy, rapidly draining battery-operated devices.

Edge devices, however, are highly heterogeneous and often feature limited memory budgets and computational power.
Optimizing ML inference to run on such devices has therefore been studied in-depth in recent years, often featuring limited prediction quality drops for significant reductions in resource usage~\cite{edgegood}.
Such approaches can be grouped in two main families: \textit{static} and \textit{dynamic}~\cite{adaptivereview}.
Static techniques include popular approaches such as quantization~\cite{quantization} and pruning~\cite{sparsity} and consists of changing the model either at training time or post-training.
Their main limitation lies in their static nature, as they are unable to adapt to changes in the external conditions, such as low battery or thermal throttling.
Dynamic inference techniques address this limitation by reconfiguring the computational graph based on the input sample complexity.
Samples that are deemed ``easy'' (e.g., a clear and centered subject in an image classification task) require only a subset of the computational budget to be correctly labeled.
On the other hand, ``complex'' inputs (e.g., a blurry image) require the full compute budget to avoid misclassification.
The core of dynamic inference is the \textit{discrimination policy} used to determine the input complexity, as it controls the accuracy vs latency tradeoff.
A conservative decision improves the accuracy, but increases the energy and latency cost. On the other hand, an underestimation may lead to a misclassification and thus to a degradation of the prediction quality. Moreover, the discrimination policy must be lightweight to avoid overshadowing the energy gains from the reduced computations.

A standard example of dynamic inference is the \textit{Big-Little}~\cite{biglittle}, where the cascade of models consists of two models: an inexpensive model (Little) and an expensive but more accurate model (Big).

\begin{algorithm}[t]
\caption{Big-Little inference}
\label{alg:big_little_inference}
\begin{algorithmic}[1]
\STATE \textbf{Input:} sample $x$, Little model $f_L$, Big model $f_B$, threshold $\tau$
\STATE $p \leftarrow f_L(x)$ \hfill (class probabilities)
\IF{$\max(p) \ge \tau$}
    \STATE $\hat{y} \leftarrow \arg\max\, p$ 
\ELSE
    \STATE $\hat{y} \leftarrow \arg\max f_B(x)$
\ENDIF
\STATE \textbf{Return:} $\hat{y}$
\end{algorithmic}
\end{algorithm}

Algorithm~\ref{alg:big_little_inference} reports an overview of the inference mechanism, shown in a more general form in Figure~\ref{fig:dynamic_inference_intro}.
The Little model is evaluated first on the input, and its output probabilities are used to decide whether to invoke, on the same input, the Big model.
The discrimination policy in this case is named $Max$ and compares the largest probability with a user-defined threshold $\tau$.
As long as most inputs require only the Little model, significant energy and average latency savings can be achieved.
On the other hand, the worst case latency of the inference increases, as it now includes an execution of the Little and Big model. Real-time systems have to consider this additional complexity layer when leveraging dynamic inference.
Our work addresses the variability in the resources and in the data distributions, balancing energy efficiency with latency deadlines.

%% file: 03_related.tex
\section{Related Work}~\label{sec:related}
Several works have explored dynamic inference to improve the energy-accuracy tradeoff.
Early-exit networks add intermediate classifiers inside the backbone model, sharing a subset of layers and introducing one or more ``exit'' points in the computational graph.
This can reduce computation by reusing intermediate activations, but may lead to sharper accuracy drops~\cite{adaptivereview}.
BranchyNet~\cite{branchynet} augments CNNs with multiple exit branches and uses the entropy of the softmax output as a confidence estimate to decide whether to stop early. This approach yields speedups ranging from 2x to 6x on tasks running on CPUs and GPUs.
The authors of~\cite{early_exit_with_latency} select exit positions during training to satisfy latency budgets, yielding
0.87\% accuracy improvements with a 37.14\% average latency reduction. At runtime, thresholds ($\tau$) are lowered when latency constraints are violated.
In~\cite{tan2021empowering}, the authors propose an optimization algorithm to select the optimal threshold values on a validation set, given a set of latency budgets.
Aside from early-exit models, cascade of classifiers such as the Big-Little~\cite{biglittle} have widely been explored in the literature.
In~\cite{wang2017idk}, the authors propose cascades of models with increasing computational cost and accuracy, reporting up to 37\% latency reductions without accuracy drops on CIFAR-10.
The authors of~\cite{cavigellicascade} analyze cascades of increasingly complex classifiers and study the effects of threshold selection and stopping policies, showing 3.1x latency improvements at iso-accuracy and evaluating cascades with mild data distribution shifts.

Differently from early-exits, our adaptive cascade is not fixed after deployment and can be reconfigured and updated dynamically to adjust to changes in external conditions.
Moreover, it adds no complexity to training, since SPs can be trained independently without architectural modifications for parameter sharing.
Our work is orthogonal to threshold selection approaches.
Instead of only tuning thresholds, we adapt at the granularity of models (e.g., enabling/disabling SPs) because, on unseen data, lowering a threshold does not necessarily preserve a hard worst-case latency constraint.

%% file: 04_methodology.tex
\section{Methodology}~\label{sec:methodology}
\begin{figure}
    \centering
    \includegraphics[width=0.55\linewidth]{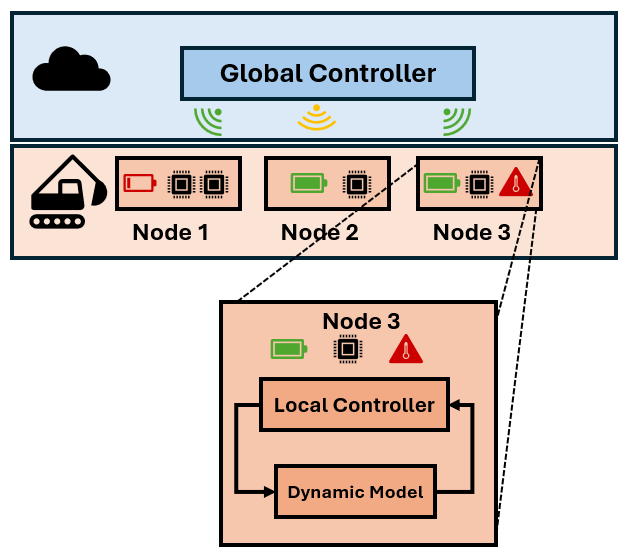}
    \caption{Dynamic Inference-aware architecture overview. 
    }
    \label{fig:software_architecture}
\end{figure}

Figure~\ref{fig:software_architecture} shows an overview of our hierarchical architecture for dynamic inference.
A fast on-device \textit{Local Controller} (LC) adapts the \emph{active} cascade at inference-time granularity, while a slower \textit{Global Controller} (GC) computes (and, when needed, updates) a budget-feasible configuration and can deploy additional SPs.
At initialization, the GC selects, for each node (hosting one task in our setting), a cascade configuration that balances accuracy and average latency under memory and worst-case latency constraints.
At runtime, the LC uses on-device measurements to react to distribution shifts and resource variability without requiring continuous communication with the GC.
The following subsections detail the dynamic inference mechanism and the two controllers.

\subsection{Dynamic Inference}\label{subsec:methodology_dynamic}
\begin{figure}[ht]
    \centering
    \includegraphics[width=0.7\linewidth]{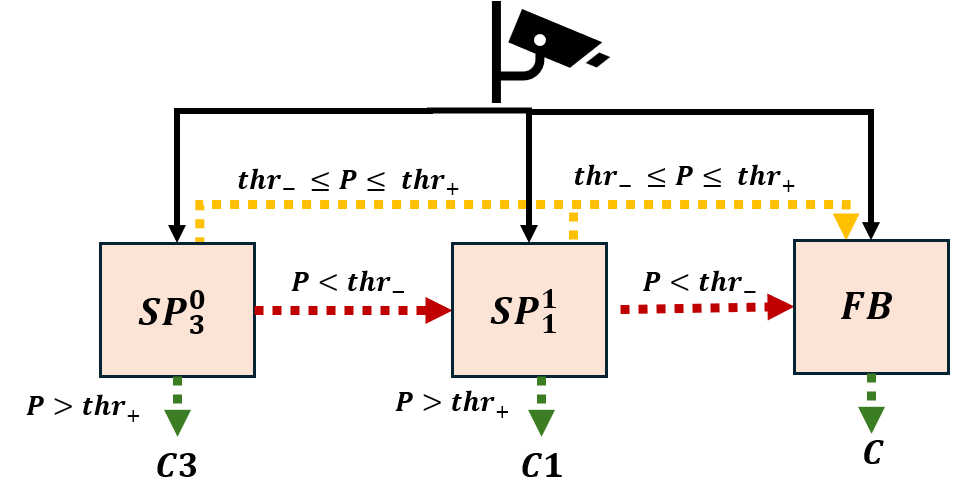}
    \caption{Dynamic cascaded inference with four classes and two SPs. 
    }
    \label{fig:dynamic_inference}
\end{figure}
We decompose the multiclass problem of our classification tasks into multiple sub-tasks.
Given a set of labels $C=\{C_1,\dots,C_N\}$, we train $N$ specialized predictors ($SP_c$), each with the purpose of estimating whether an input $x$ belongs to class $c$. 
Due to the lower complexity of the subtask, these binary models are significantly smaller than the full multiclass model in both memory and computational cost and can be designed independently depending on the specific label complexity.
In this work, we keep the SP architecture identical for a task, leaving sub-task-specific architecture as future work.

At task scheduling time, the GC evaluates a subset ($D$) of these predictors in a predefined order ($o$) for each task $t$, that we represent as follows:
\begin{equation}
    O_t = \{SP^{0},\dots,SP^{D},FB\}
\end{equation}
with $D\leq|C|$ and $FB$ being the original multiclass model.
This ordering is selected according to the deployment and latency constraints, as we detail in Section~\ref{subsec:global_scheduling}.

At inference time, the cascade of the deployed models is evaluated as follows. Each $SP^i$ employs two thresholds $thr_{-}$ and $thr_{+}$ to determine its decision:
\begin{equation}~\label{eq:dynamic_inference}
    d_{i}(x) = 
      \begin{cases}
    \! %
    \begin{alignedat}{2}
      & P^i_c(x)>thr_{+} && \text{Accept}\ c 
      \\
      & P^i_c(x)<thr_{-} && \text{Reject}\ c
      \\
      & thr_{-} \leq P^i_c(x) \leq thr_{+}\ && \text{Undecided}\ c
    \end{alignedat}
  \end{cases}
\end{equation}
where $P^i_c(x)$ is the predicted probability by the SP in the cascade at position $i$ for the for class $c$.
An accepted decision ends the inference early.
A reject forces the execution of the following $SP$, or eventually the multiclass model $FB$.
Intuitively, in both of these cases, the classifier is confident enough in its prediction that $x$ does or does not belong to class $c$.
If undecided, i.e., we are close to a random prediction of 0.5, we invoke FB.
The intuition is that the other SPs in the pipeline are unlikely to give highly confident and correct probabilities, and therefore executing them is suboptimal.

This inference pipeline has an expected inference cost ($E_{inf}$) in multiply-and-accumulate (MAC) operations of:
\begin{equation}
E_{\text{inf}}=\sum_{i=0}^{D} \mathbb{1}{(\text{$SP^{i})$}}*E_{SP^{i}} + \mathbb{1}{(\text{$FB$})}*E_{FB}
\end{equation}
where $E_{SP^i}$ and $E_{FB}$ are respectively the computational costs of $SP^i$ and $FB$ and $\mathbb{1}{(X)}$ is 1 if model $X$ has been executed.

Due to the inexpensive nature of each SP, this approach leads to consistent savings as long as the FB is rarely invoked.
However, SPs are generally not as accurate as the FB; as a consequence, increasing $D$ trades off accuracy for higher energy efficiency and lower average latency.
Conversely, the worst case latency increases together with $D$, as a real-time system must consider a full invocation of the inference pipeline as the upper bound.
In terms of memory, the overhead introduced due to storing multiple predictors is generally negligible, as these binary classifiers are orders of magnitude smaller than the FB model.
Finally, $thr_-$ and $thr_+$ are selected empirically depending on the task, possibly leveraging heuristic-based approaches that are orthogonal to this work.
Figure~\ref{fig:dynamic_inference} shows an example with $C=4$ and $D=2$, where only the $SP$ for classes 3 and 1 are deployed.
Following Equation~\ref{eq:dynamic_inference}, the input traverses the cascade until an early exit is triggered or the FB model is run.
This design is flexible: small one-vs-rest SPs can be enabled or disabled on demand, unlike heavyweight multiclass models in a fixed cascade, adapting to increase in the timing pressure due to change in the available resources.
Finally, under distribution shift, the cascade can also be re-ordered, continuously adapting to maintain high energy efficiency and low average latency by maximizing the early exits.

\begin{figure*}[ht]
    \centering
    \includegraphics[width=0.55\textwidth]{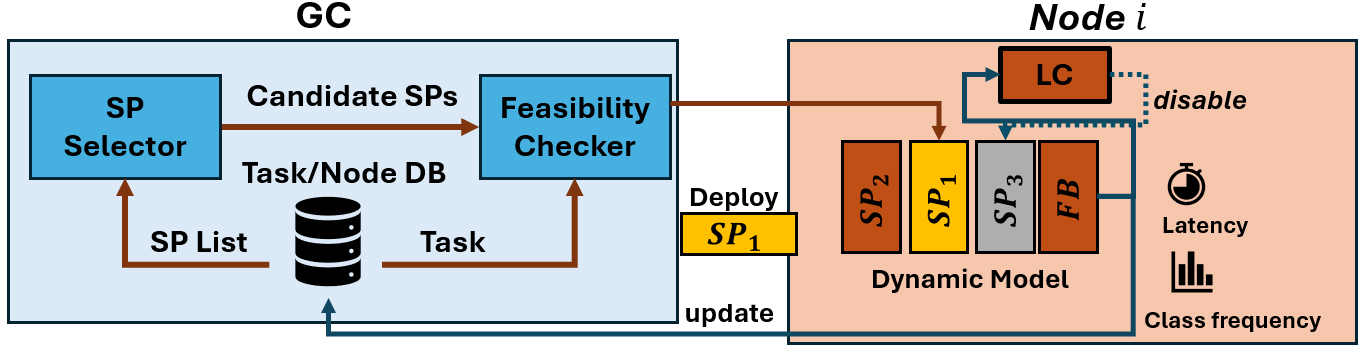}
    \caption{Framework runtime overview.}
    \label{fig:framework_runtime}
\end{figure*}
\subsection{Global Controller}~\label{subsec:global_scheduling}
The GC implements two core functions in the proposed architecture, providing both the initial task scheduling and online adaptation.

\subsubsection{Task Scheduling and Model deployment}
In our setting, each edge node hosts at most one ML task, and the role of the GC is to select a node-feasible dynamic-inference configuration. 
The GC follows a two-stage, heuristic selection procedure.

First, it builds a set of candidate SP cascades for a task using a beam search that retains only a limited number of high-quality orders per depth, minimizing a score based on the expected MACs per inference (Equation~\ref{eq:dynamic_inference}) on the validation set. 
Second, it filters out cascades that exceed a user-defined accuracy drop w.r.t. the fallback model and removes configurations that violate node-level memory constraints or the task's worst-case latency constraint.
Finally, among the remaining feasible candidates, the GC selects the configuration with the minimum score and deploys the corresponding set and ordering of SPs.

\subsubsection{Online Update}

At runtime, the GC receives updates from each node, tracking changes in hardware conditions (e.g., temperature and battery lifetime) and task-level signals (e.g., predicted class frequency), as shown in Figure~\ref{fig:framework_runtime}.

The SP selector takes the task SPs from the local storage and determines if deploying a new SP can improve the energy efficiency while maintaining an acceptable accuracy drop.
For instance, in the case of class distributions diverging from the validation set, it focuses first on the SPs trained on high-frequency classes.
Once a set of candidate SPs has been found, the Feasibility Checker verifies that adding a model to the cascade does not violate the node and latency constraints.

In case adding an SP is deemed feasible, the GC transmits the new SP to the node, where it is stored for future use.
Notably, this approach delays a forced full reconfiguration of the cascade and, as long as it is performed with low frequency, it is lightweight on the node's battery.

\subsection{Local Controller}

The LC, shown in Figure~\ref{fig:framework_runtime}, monitors node-level resources, such as CPU availability, memory, and temperature.
It also keeps statistics of the task deployed on the same node, with online updates after each inference.
Specifically, the LC computes the exponential moving average (EMA) of the predicted classes, providing an updated estimate of the predicted class distribution, and is used by the GC and LC to reconfigure the $SP$ at runtime.

The LC can perform three actions.
First, it can disable an SP if the corresponding class frequency is lower than a threshold~$\tau$.
Second, it can re-enable a locally available SP (i.e., it was either deployed at scheduling time or sent later by the GC) if its class frequency is larger than~$\tau$.
Third, the LC sorts the enabled SPs according to the EMA.

To provide timing robustness, the LC enforces a simple per-inference safety check.
Given $D_t$ and $L_k$ denoting respectively the latency constraint of task $t$, and the end-to-end inference latency of the sample $k$ (including the LC status update), if $L_k > D_t$, the LC reduces the active cascade (e.g., disabling the last enabled SPs) until the worst-case latency of the active configuration fits within $D_t$.
In practice, we reserve a fixed budget $T_{LC}$ for LC actuation and enforce:
\begin{equation}
T_{LC} + \sum_{SP\in\mathcal{A}} T(SP) + T(FB) \leq D_t,
\end{equation}
where $\mathcal{A}$ is the set of enabled SPs.
This mechanism ensures that adaptation-induced latency spikes remain bounded by design, while the EMA-driven logic preserves efficiency under distribution mismatch.

In general, SPs may be aggressively enabled in low-battery mode, as they reduce the average energy per inference, while the sorting mechanism is used to make the system more resilient to test-time class distribution drifts.
Intuitively, an SP executed on a class that rarely appears due to class drifts wastes computations and energy, while also increasing the possibility of deadline misses.
Therefore, our EMA-based control will first move such SP towards the end of the cascade, eventually switching it off entirely.

%% file: 06_experimental_evaluation.tex
\section{Results}~\label{sec:results}
\vspace{-8mm}

\begin{table*}[h]\caption{Deployment results}\label{tab:deployment}
\centering
\begin{tabular}{cllrrrrrr}
\multicolumn{1}{l}{} &
   &
   &
  \multicolumn{2}{c}{\textbf{Base}} &
  \multicolumn{2}{c}{\textbf{Mismatch  Minor}} &
  \multicolumn{2}{c}{\textbf{Mismatch  Major}} \\ \hline\hline
\multicolumn{1}{l}{\textbf{Dataset}} &
  \textbf{Method} &
  \textbf{SP Order} &
  \textbf{Latency{[}ms{]}} &
  \textbf{Energy{[}mJ{]}} &
  \textbf{Latency{[}ms{]}} &
  \textbf{Energy{[}mJ{]}} &
  \textbf{Latency{[}ms{]}} &
  \textbf{Energy{[}mJ{]}} \\ \hline
\multirow{3}{*}{SCD}     & Static  &         & 14.24 & 18.13 & 14.24 & 18.13 & 14.24 & 18.13 \\
                         & Dynamic & 6, 2, 5 & 5.66  & 6.31  & 4.05  & 4.07  & 9.81  & 11.81 \\
                         & Ours    & 6, 2, 5 & 5.82  & 6.32  & 3.98  & 3.99  & 5.96  & 6.54  \\\hline
\multirow{3}{*}{CIFAR-10} & Static  &         & 33.65 & 43.88 & 33.65 & 43.88 & 33.65 & 43.88 \\
                         & Dynamic & 3, 4    & 30.35 & 38.62 & 27.41 & 34.91 & 30.14 & 38.62 \\
                         & Ours    & 3, 4    & 28.32 & 36.28 & 24.22 & 30.65 & 28.28 & 36.28
\end{tabular}
\end{table*}

\subsection{Experimental Setup}

We deploy the LC and dynamic models on a Raspberry Pi 4B, and run the GC on an Asus NUC (quad-core Intel i7, 16 GB RAM).
We measure Raspberry Pi energy consumption with an inline USB power meter.
We evaluate on the Soil Classification Dataset (SCD)~\cite{soil_dataset} and CIFAR-10~\cite{cifar10}.
For SCD (5097 images, 7 classes, imbalanced), we use an 80/10/10\% train/val/test split, while for CIFAR-1.0, we split the training set into 80/20\% train/val and use the standard test set.
We enforce per-inference latency constraints of 0.05\,s for SCD and 0.1\,s for CIFAR-10.

\begin{table}[ht]\caption{Baseline static models}~\label{tab:baseline}
\resizebox{\linewidth}{!}{%
\begin{tabular}{cllll}
\multicolumn{1}{l}{\textbf{Dataset}} & \textbf{Model} & \textbf{MAC{[}M{]}} & \textbf{Params{[}K{]}} & \textbf{B.Acc{[}\%{]}} \\ \hline \hline
\multirow{2}{*}{SCD}     & SP               & 0.65  & 61.24   & 80.8 \\
                         & MobileNetV2 w0.5 & 31.64 & 696.65  & 90.5 \\ \hline
\multirow{2}{*}{CIFAR-10} & SP               & 0.65  & 61.24   & 74.5  \\
                         & MobileNetV2 w1   & 94.54 & 2236.68 & 90.8 
\end{tabular}
}
\end{table}

SPs are LeNet binary classifiers, while FB models are MobileNetV2 (width 0.5 for SCD, 1 for CIFAR-10).
Table~\ref{tab:baseline} reports the models' parameters and MACs.
We train all models with PyTorch 2.5 on the NUC and run the inference with ONNX Runtime on the Raspberry Pi. Due to the class imbalance of the tasks, we report the results using the balanced accuracy.

\subsection{GC Scheduling Trade-offs}
\begin{figure}[h]
\centering
\includegraphics[width=0.55\linewidth]{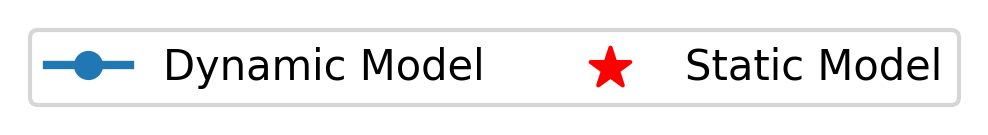}
\begin{subfigure}{0.45\linewidth}
\includegraphics[width=\linewidth]{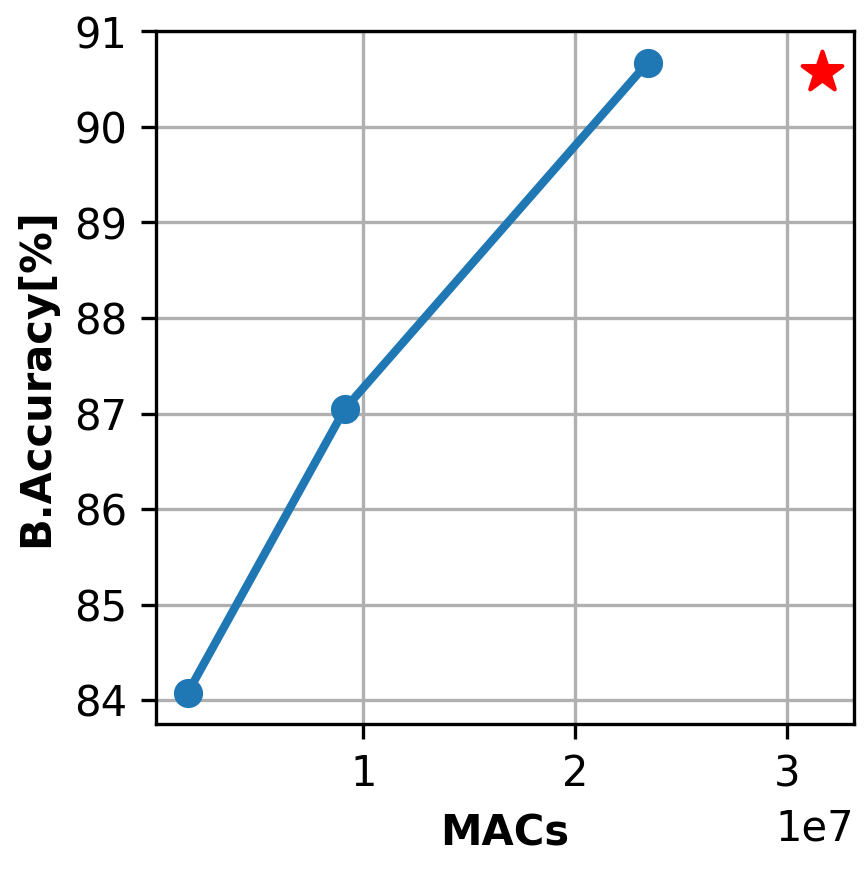}
\subcaption{SCD}
\end{subfigure}
\begin{subfigure}{0.45\linewidth}
\includegraphics[width=\linewidth]{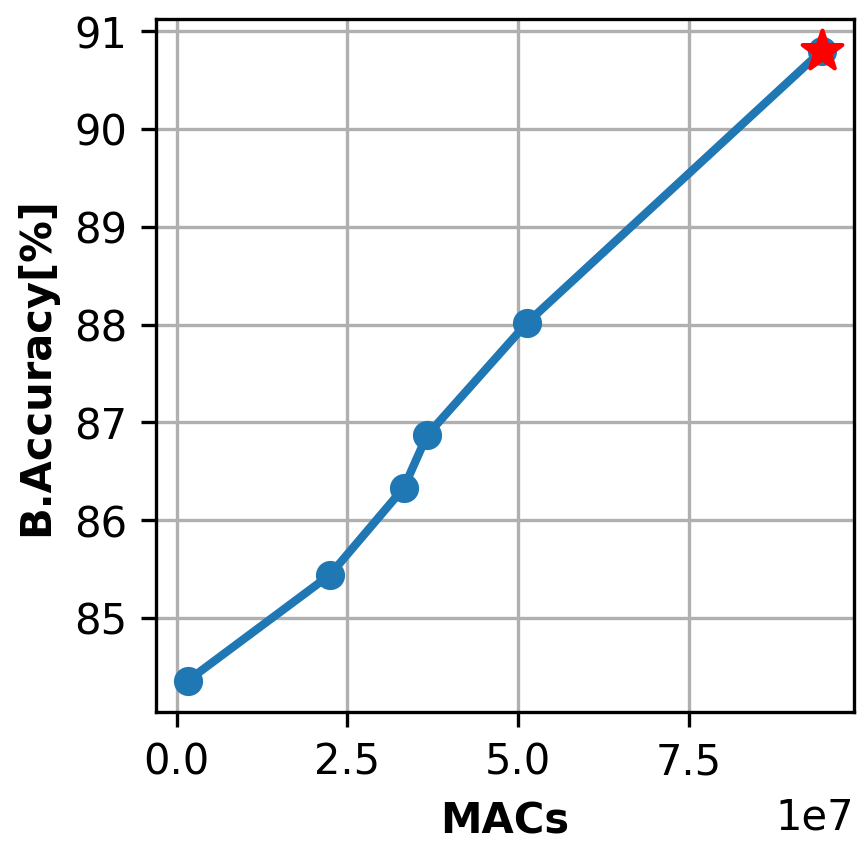}
\subcaption{CIFAR-10}
\end{subfigure}
\caption{GC scheduling results at different accuracy versus MAC tradeoffs.}
\label{fig:dynamic_inference_results}
\end{figure}

Figure~\ref{fig:dynamic_inference_results} reports the Pareto frontier between accuracy and compute (MACs), which we use to select runtime configurations that satisfy the latency constraints while minimizing on-device cost.
Following the GC procedure in Section~\ref{subsec:global_scheduling}, we generate candidate cascades, prune those violating the memory/worst-case latency constraints or exceeding the target accuracy drop, and select the feasible order that minimizes the expected MACs.
We repeat this process for each target accuracy drop in $[0, 10]\%$ (step 1\%), obtaining the \textit{Dynamic Model} points. The FB model (red) provides the static reference point, i.e., the MobileNet architectures in Table~\ref{tab:baseline}.

Concerning SCD, we obtain an increase in balanced accuracy of 0.1\% with a reduction of MACs of 25.98\% w.r.t. FB. 
The binary classifier for the yellow soil becomes even more accurate than the FB model on its target class, causing the cascade of two models to become more accurate than the single static one.
If we allow an accuracy drop of 3.52\%, the MAC reduction we obtain is 94.57\%.

Concerning CIFAR-10, the accuracy versus MAC curve is steeper due to the higher task complexity.
A drop of 2.78\% accuracy reduces the MACs by up to 45.66\%, while a drop of 6.44\% yields savings of 98.28\% w.r.t. the FB.

\subsection{Deployment Results}

For the on-device experiments, we select the most efficient configuration in Figure~\ref{fig:dynamic_inference_results} that introduces at most a 4\% accuracy drop with respect to the FB model.
These orders allow us to benchmark ordering with $D>1$, where our LC and GC can fully apply optimizations such as SP sorting.
Moreover, to emulate an actual continuous data stream, we iterate over each test set three times.

All workloads execute inside a Docker container, limited to a single CPU core for easier comparisons.
With this compute budget, the execution time of a single SP is 0.44 ms.

We evaluate three scenarios that capture stationary conditions (\textbf{Base}) and distribution shifts that change the frequency of the early-exit classes (\textbf{Mismatch Minor/Major}).
Table~\ref{tab:deployment} reports average latency and energy per inference when running only the FB model (\textit{Static}), a dynamic cascade without LC and GC adaptation (\textit{Dynamic}) and our approach (\textit{Ours}).

\textbf{Base} uses the original test set as a steady-state reference.
For SCD, our approach reduces average latency by 2.45$\times$ vs. the static configuration, while non-adaptive dynamic inference achieves 2.52$\times$.
Energy follows the same trend. Our approach and dynamic inference yield up to 2.86$\times$ and 2.87$\times$ savings vs. static.
For CIFAR-10, our approach improves latency by 1.18$\times$ and 1.07$\times$ over static and dynamic baselines, respectively. It reduces energy by up to 1.20$\times$ vs. static.

The mismatch behavior differs across datasets.
CIFAR-10 has a uniform class frequency and a larger test set, so our solution can update the moving average and adapt the cascade order.
For SCD, the deployed order already includes the three most frequent classes in decreasing order.
No changes are then needed at runtime, as the order is already optimal for the label distribution.
The remaining overhead introduced by LC and GC is mainly due to their Python implementation, while inference relies on the optimized ONNX Runtime.

\textbf{Mismatch Minor} augments by five times the samples of the class corresponding to the last SP in the cascade.
This is a corner case for non-adaptive dynamic inference, where the efficiency of the deployed order diminishes.
For SCD, we obtain a latency reduction of 3.57$\times$ and energy reduction of 4.54$\times$ w.r.t. static.
We slightly outperform dynamic inference, reducing the average latency by 0.07~ms and energy by 0.08~mJ.
For CIFAR-10, we reduce latency by 1.39$\times$ and energy by 1.43$\times$ vs. static.
We also outperform dynamic inference, with latency and energy reductions of 1.13$\times$.

\textbf{Mismatch Major} replicates by the same factor a class not included in the deployed order (class 0 in our experiment).
For SCD, our solution yields a speedup of 2.38$\times$ and an energy reduction of 2.77$\times$ w.r.t. static.
Dynamic inference deteriorates, yielding lower savings in latency (1.45$\times$) and energy (1.53$\times$).
For CIFAR-10, we achieve a speedup of 1.19$\times$ and an energy reduction of 1.21$\times$ vs. static.

Overall, stronger test-time divergence increases the benefit of closed-loop adaptation.
Our framework is the only method among the baselines that can adjust the deployed cascade online to recover efficiency under distribution shift.

\subsection{Self-Adaptation Analysis}
\begin{figure}[ht]
    \centering
    \includegraphics[width=0.9\linewidth]{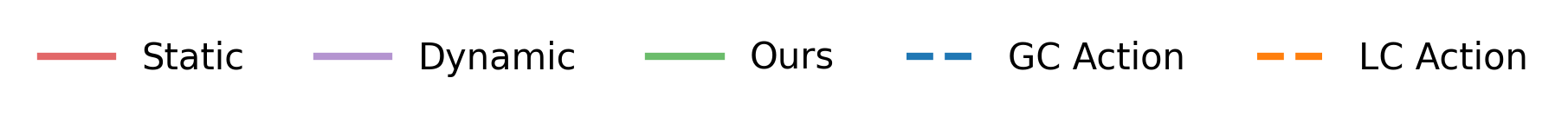}
    \includegraphics[width=0.75\linewidth]{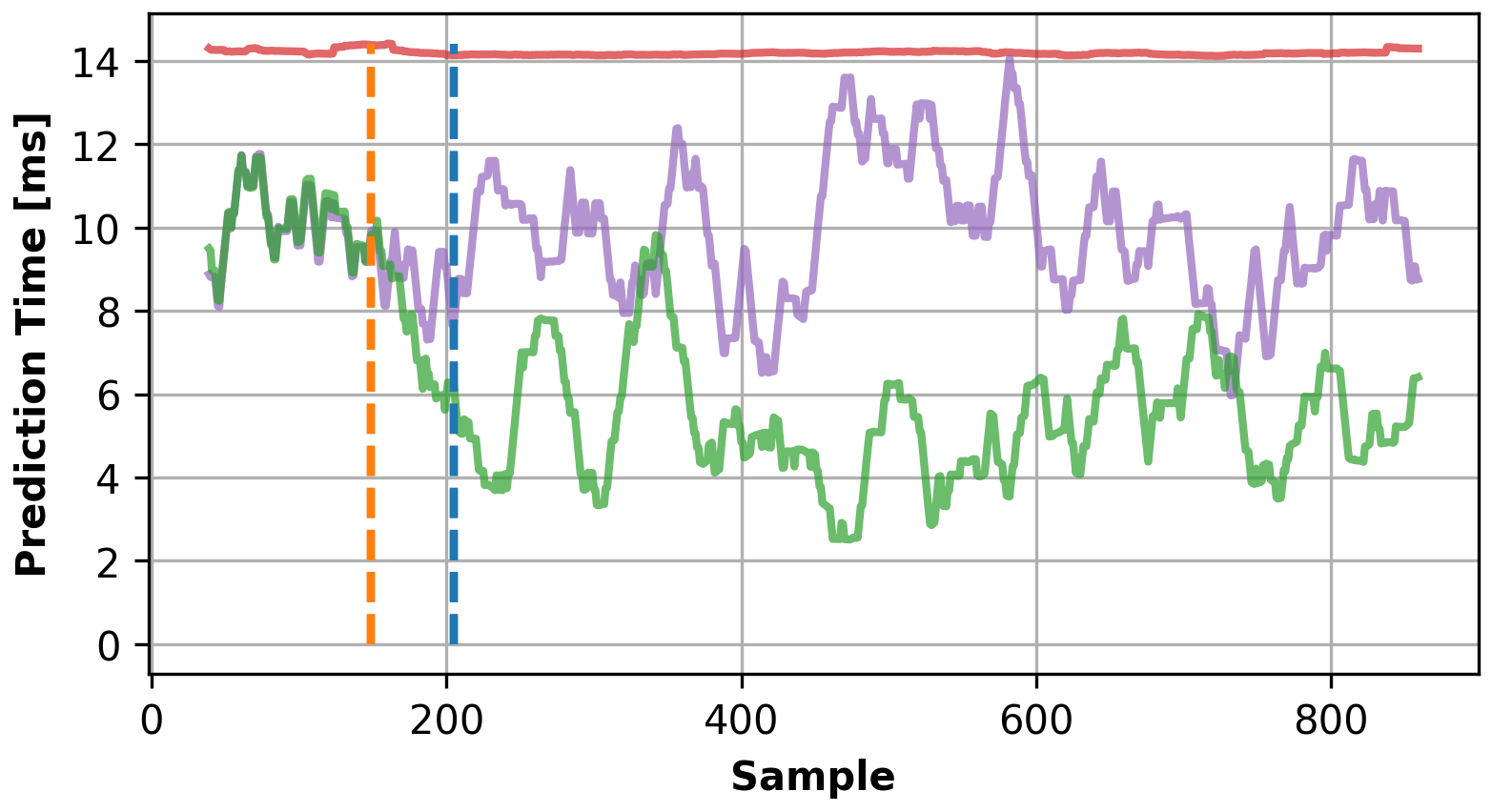}
    \caption{SCD Mismatch Major inference overview.}
    \label{fig:detail_result}
\end{figure}

Figure~\ref{fig:detail_result} shows the first iteration on the test of the SCD Mismatch Major scenario detailed in Table~\ref{tab:deployment}.
We report the latency per inference with a moving average smoothing for easier readability.

As expected, the Static baseline remains constant over the entire duration at approximately 14 ms.
The classic dynamic solution reduces the average latency, but it suffers from different class distributions.
It also causes sharp peaks in inference latency, matching a static solution after 600 samples.

Our framework fully handles the mismatch.
After 180 inferences, the LC (orange vertical line) detects a significant change in the class distribution and disables the least invoked SP in the deployed order to reduce the average inference latency.
At 200 inferences, our GC (blue vertical line) is triggered and offloads the SP corresponding to the augmented class 0 to the node.
Across the entire execution, our framework performs consistently better or close to the standard dynamic inference approach.
At the node level, for the Mismatch Major scenario, we reduce the total energy used by the Raspberry Pi by 58\% with respect to a static solution and by 39\% with respect to the Dynamic baseline.

%% file: conclusions.tex
\section{Conclusion}~\label{sec:conclusion}
\vspace{-2mm}

This work introduces an adaptive software architecture for dynamic inference based on a hierarchical control loop.
A global controller (GC) and local controller (LC) jointly adapt the deployed model configuration to satisfy resource and latency constraints while reducing the average latency and energy per inference.

We validate the approach on two vision workloads (SCD and CIFAR-10) and show that dynamic adaptation can substantially improve efficiency with limited accuracy loss compared to static baselines.

Future work will extend the controllers to additional drift sources (e.g., lighting changes or sensor degradation) and broader model families.

\section*{Acknowledgement}
Supported by Next-Generation AI (Forskningsreserven). This work is supported by Digital Research Centre Denmark (DIREC).